%% file: main.tex
\documentclass[final]{cvpr}

\usepackage{times}
\usepackage{epsfig}
\usepackage{graphicx}
\usepackage{amsmath}
\usepackage{amssymb}
\usepackage[dvipsnames]{xcolor}
\usepackage[font={small}]{caption}
\usepackage{verbatim}
\usepackage{multirow}
\usepackage{booktabs} % for pretty plots
\usepackage{subfig}
\usepackage{color}

\usepackage[pagebackref=true,breaklinks=true,colorlinks,bookmarks=false]{hyperref}

\def\cvprPaperID{****} % *** Enter the CVPR Paper ID here
\def\confYear{CVPR 2021}

\begin{document}

%%%%%%%%% TITLE
\title{Human Mesh Recovery from Multiple Shots}

\author{Georgios Pavlakos, Jitendra Malik, Angjoo Kanazawa \\[0ex]
University of California, Berkeley\\
%{\tt\small \{pavlakos,malik,kanazawa\}@berkeley.edu}
}

\twocolumn[{%
\renewcommand\twocolumn[1][]{#1}%
\maketitle
\begin{center}
\newcommand{\teaserwidth}{\textwidth}
\vspace{-0.2in}
\centerline{
\includegraphics[width=\teaserwidth,trim={0 8cm 18.5cm 0},clip]{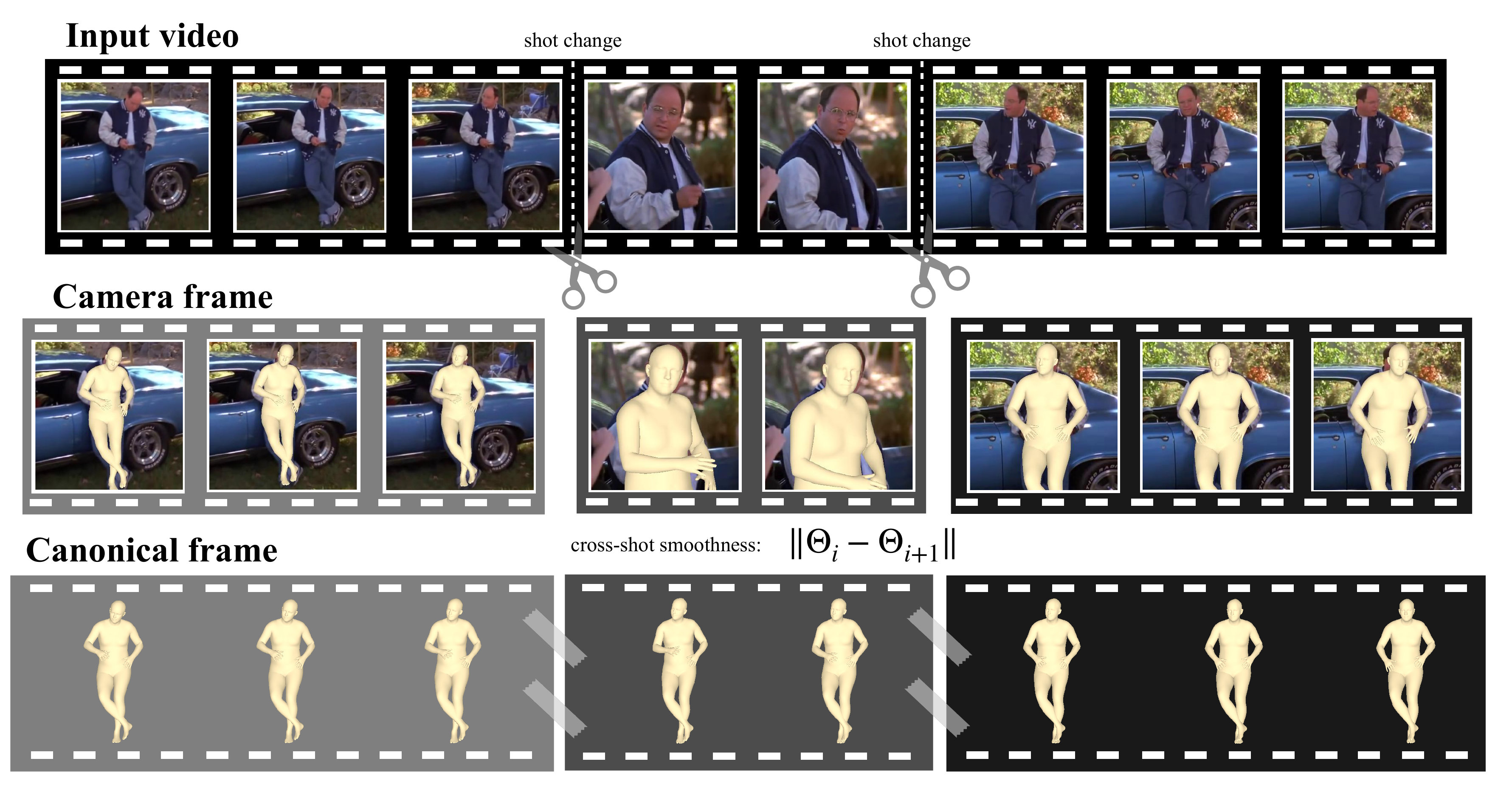}
 }
\vspace{-0.1in}
\captionof{figure}{\small{\bf Human Mesh Recovery from Multiple Shots.} Videos from edited media, like movies, include sudden shot changes that lead to discontinuities between the frames (top), which reduce  the  rich  potential  of  a  film to a series of short independent temporal sequences.  However within the same scene, the underlying 4D structure of the scene only changes smoothly.  We leverage this insight and treat the different shots as multi-view cues that provide complementary information about the 3D human body underlying these shot boundaries. This leads to both more accurate 3D reconstructions (middle, bottom) and longer 3D pose sequences, which prove to be a great source of data for training deep learning models.}
%\vspace{-0.15in}
\vspace{-1em}
\label{fig:teaser}
\end{center}
}]
\maketitle

\begin{figure*}[h]
\centering
\includegraphics[width=\textwidth,trim={0 18.5cm 24.5cm 0},clip]{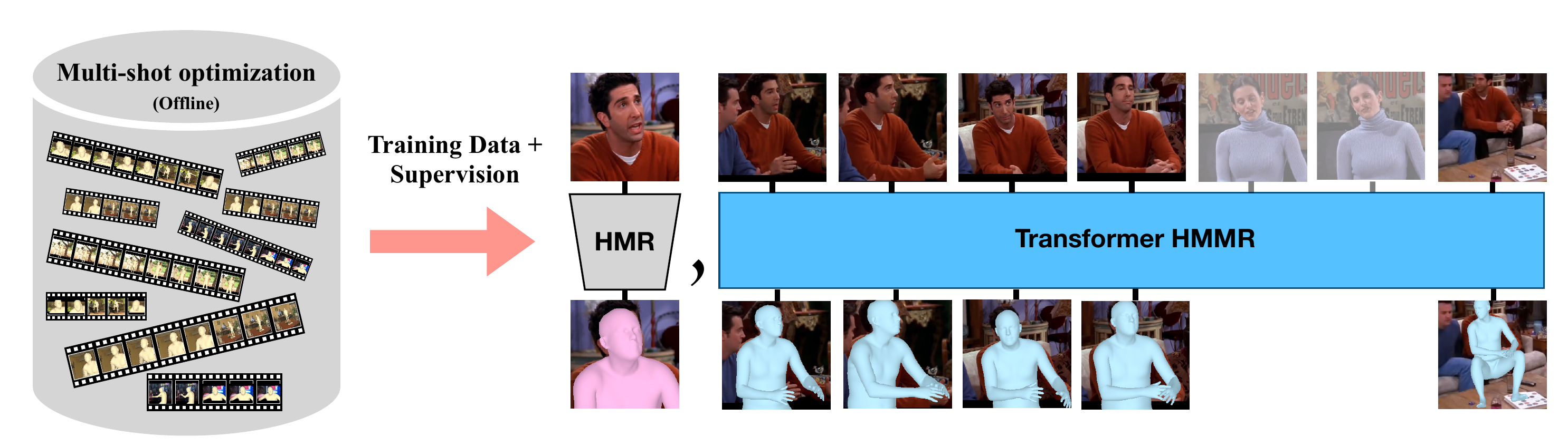}
\caption{{\bf Overview of our workflow.} We reconstruct 3D human mesh sequences from movies using multi-shot optimization. The resulting reconstructions can be used as training data for both single-view human mesh recovery and temporal human mesh motion recovery via the proposed transformer-based model. 
}
%\vspace{-.8em}
\label{fig:overview}
\end{figure*}

%%%%%%%%% ABSTRACT
\input{01_abstract.tex}

\section{Introduction}
\input{02_introduction.tex}

\section{Background}
\input{03_background.tex}

\section{Multi-shot optimization}
\input{04_multi_shot.tex}

\section{Human Mesh Recovery}
\input{05_human_mesh_recovery.tex}

\section{Experiments}
\input{06_experiments.tex}

\section{Conclusion}
\input{07_conclusion.tex}

\color{black}
{\small
\bibliographystyle{ieee_fullname}
\bibliography{egbib}
}

\end{document}

%% file: 01_abstract.tex
\begin{abstract}
\vspace{-0.8em}
Videos from edited media like movies are a useful, yet under-explored source of information. The rich variety of appearance and interactions between humans depicted over a large temporal context in these films could be a valuable source of data. However, the richness of data comes at the expense of fundamental challenges such as abrupt shot changes and close up shots of actors with heavy truncation, which limits the applicability of existing human 3D understanding methods. In this paper, we address these limitations with an insight that while shot changes of the same scene incur a discontinuity between frames, the 3D structure of the scene still changes smoothly. This allows us to handle frames before and after the shot change as multi-view signal that provide strong cues to recover the 3D state of the actors. We propose a multi-shot optimization framework, which leads to improved 3D reconstruction and mining of long sequences with pseudo ground truth 3D human mesh. We show that the resulting data is beneficial in the training of various human mesh recovery models: for single image, we achieve improved robustness;  for video we propose a pure transformer-based temporal encoder, which can naturally handle missing observations due to shot changes in the input frames. We demonstrate the importance of the insight and proposed models through extensive experiments. The tools we develop open the door to processing and analyzing in 3D content from a large library of edited media, which could be helpful for many downstream applications. Project page: \url{https://geopavlakos.github.io/multishot}
\vspace{-3.8em}
\end{abstract}

%% file: 02_introduction.tex
Movies are a treasure trove of human ``behavior episodes'' \cite{barker1955midwest}. They are produced in many different countries  in multiple genres,  giving us tremendous cultural diversity and range. Datasets,  most prominently,   AVA \cite{gu2018ava} have emerged which provide a rich annotation of spatio-temporally localized human actions in movies. This would seem like ideal data on which to train systems for video understanding, and furthermore use that as a stepping stone for acquiring ``common sense'' from observations of diverse human behavior. This ``visual'' route could be complementary to the ``linguistic'' route to capturing common sense and arguably more fundamental.

But before we go too far with our wishful thinking,  we must confront a fundamental challenge of video data derived from movies – the complication of “shots”. Film has a grammar~\cite{arijon1976grammar}. Stories are communicated through a juxtaposition of shots, typically from different camera angles viewing the same scene. Alfred Hitchcock’s Rope and Sam Mendes’s 1917 are noteworthy precisely because they are presented as a single take, without any discernible breaks corresponding to shot boundaries.

These  shot  changes  manifest  as  sudden  discontinuities  in video as illustrated in Figure~\ref{fig:teaser}.  Current temporal 3D human mesh and motion recovery methods, as well as most action classification algorithms, treat these shots as independent scenes, which reduce the rich potential of a film to a series of short independent temporal sequences.  We see this as a lost opportunity,  as these shots depict a single underlying 4D scene seen from different viewpoints despite the temporal discontinuities at the frame level. Properly modeled shot changes can thus be  powerful signal rather than noise, as they provide a multi-view signal of the underlying dynamic scene. This can be a powerful cue in disambiguating the underlying the 3D pose and motion of humans,  which  is  particularly  helpful  for  close-up,  heavily truncated image of people.   We take this insight and propose a multi-shot optimization procedure,  which recovers a consistent 3D motion sequence across shot changes,  simultaneously addressing both challenges of temporal fragmentation and partial humans.

The proposed multi-shot optimization recovers long 3D motion sequences from movies and serve as a rich source of pseudo-ground truth 3D training data for robust human mesh recovery from images or video. This workflow is illustrated in Figure~\ref{fig:overview}.  Training a single-frame human mesh recovery model~\cite{kanazawa2018end} on the recovered 3D meshes from the multi-shot optimization results in improved robustness against heavy truncation and other ambiguities.  More importantly, on the video setting, we propose a pure transformer-based temporal human mesh and motion recovery model (t-HMMR), which we demonstrate to be not only competitive, but also a particularly suitable architecture for films. Aside from sudden shot changes, films pose a challenge that the same person may not be consecutively depicted in the scene due to shot changes to another character or a background. Transformers can easily address these situations as it can explicitly not attend to frames that do not contain the person of interest and naturally ignore them, while still processing a larger temporal context  before and after the missing input frames.

For our experiments, we employ AVA~\cite{gu2018ava}, a large scale dataset of movies with atomic action annotations. We apply our multi-shot optimization on AVA, which results in over 350k frames of data with pseudo ground truth 3D. We refer to this dataset as Multi-Shot-AVA (MS-AVA), and we use it to train regression models for human mesh recovery, both from single image (HMR) and from video (t-HMMR). We demonstrate the importance of our multi-shot optimization and the benefit of the collected data through extensive experimentation on MS-AVA and the common benchmarks.

In summary, we introduce the problem of human mesh recovery from multiple shots and we propose a novel multi-shot optimization approach. This results in a new dataset, which is used to train a more robust single frame model for human mesh recovery and a pure transformer-based temporal model. Upon publication we will release our code and MS-AVA. We hope MS-AVA opens new opportunities for many future research directions. 

%% file: 03_background.tex
This section provides reference to prior work and acts as background to our approach. The relevant literature is vast, so here we consider the most relevant approaches.

\subsection{Human body modelling}
Recent work in 3D human reconstruction has been influenced heavily by the availability of powerful human body models. The SMPL model~\cite{loper2015smpl} is one of the most popular choices that, among others, has enabled work on reconstruction~\cite{kanazawa2018end}, prediction~\cite{zhang2019predicting}, as well as imitation~\cite{peng2018sfv}. At a high level, one can consider SMPL as a function $\mathcal{M}(\theta, \beta)$ that takes as input pose parameters $\theta$ and shape parameters~$\beta$ (collectively $\Theta=\{\theta, \beta\}$) and returns the 3D body mesh $M$ and joints $X$.  Other body models follow similar formulations, with differences on the modelling side~\cite{haoyang2020blsm,osman2020star,xu2020ghum}, or the expressivity of the model~\cite{anguelov2005scape,joo2018total,pavlakos2019expressive}.

\subsection{3D pose and shape from single image}
\noindent
\textbf{Optimization:}
Reconstructing 3D pose and shape from a single image is often addressed in an optimization setting. In these approaches~\cite{bogo2016keep,guan2009estimating,huang2017towards,lassner2017unite,pavlakos2019expressive,zanfir2018monocular}, a set of features are detected on an image (typically 2D keypoints), and then a configuration of the body model is recovered such that it is consistent with the features. This requires a reprojection objective $E_{\textrm{proj}}$ that penalizes deviations of the projected model from the detected features, and a set of objectives $E_{\textrm{prior}}$, that express the priors and encourage the reconstruction to be valid. At test time, the sum of these objectives is minimized in an iterative manner. The SMPLify~\cite{bogo2016keep,pavlakos2019expressive} methods are canonical examples of this type of approach for single image reconstruction,  but other settings have also been considered, \eg, from multiple views~\cite{dong2020motion,huang2017towards}, or monocular video~\cite{arnab2019exploiting,kocabas2020vibe}. In this work, we present an approach that focuses on the setting of reconstruction from \textit{multiple shots}.

\noindent
\textbf{Direct prediction:}
Directly regressing the SMPL parameters has seen many successes recently due to deep learning advances. A canonical example is HMR~\cite{kanazawa2018end}, which learns a direct mapping from raw RGB images to SMPL parameters and involves design principles adopted by many follow-up works~\cite{arnab2019exploiting,kolotouros2019learning,georgakis2020hierarchical,rong2019delving,pavlakos2019texturepose}. More specifically, HMR consist of a feature encoder $f_{\textrm{im}}: I \mapsto \phi$ that converts an image $I$ to a feature representation $\phi$, followed by an iterative feedback regressor that maps the intermediate features to model parameters, $\hat{\Theta}$, and camera parameters, $\hat{\Pi}$. Using the predicted camera parameters, the reconstructed mesh can be projected to the image, which enables supervision with reprojection losses, given 2D annotations. Concurrently with HMR, other works have investigated decoupled regression approaches~\cite{tung2017self,choi2020pose2mesh,moon2020i2l,omran2018neural,pavlakos2018learning,song2020human,xu2019denserac}, where the intermediate feature representation is hardcoded, \eg, 2D keypoints, silhouettes, semantic parts or dense correspondences.

\noindent
\textbf{Limitations:}
Previous works~\cite{joo2020exemplar,rockwell2020full} have identified the limitations of relevant reconstruction approaches particularly when it comes to heavy truncation of humans. Joo~\etal~\cite{joo2020exemplar} propose augmentation with synthetically cropped examples, while Rockwell and Fouhey~\cite{rockwell2020full}, retrain their model with confident reconstructions. In our work, we use complementary information from neighboring shots to improve the 3D reconstruction and collect training examples that improve the robustness of our HMR model.

\subsection{3D pose and shape from video}
For video approaches, the goal is 3D reconstruction given a video sequence $V = \{I_t\}^T_{t=1}$, of length $T$. Video methods that follow-up HMR, \eg,~\cite{kanazawa2019learning,kocabas2020vibe,luo20203d}, take a similar workflow with the addition of a temporal encoder function $f_{\textrm{movie}}$, which maps per-frame features $\phi_t$ to per-frame sequence features $\Phi_t$, from which the model and camera parameters for each frame are predicted via a 3D regressor $f_{\textrm{3D}}: \Phi_t \mapsto \{\hat{\Theta}_t$, $\hat{\Pi}_t\}$. These methods often differ in the choice of the architecture for the temporal encoder $f_{\textrm{movie}}$. Kanazawa~\etal~\cite{kanazawa2019learning} use a convolutional model, Kocabas~\etal~\cite{kocabas2020vibe} and Luo~\etal~\cite{luo20203d} use a recurrent model, while Sun~\etal\cite{sun2019human} use a hybrid model combining convolutions with self-attention. In our work, we propose a pure transformer model, based exclusively on self-attention. We show that transformer encoder is not only a competitive and stable architecture for video, but also a suitable architecture to handle missing identities that often occur in films. 

\subsection{Common video datasets}
Early work on 3D human reconstruction relied on motion capture datasets with full 3D supervision~\cite{sigal2010humaneva,ionescu2013human3} via mocap or multi-view stereo~\cite{mehta2017monocular}.  Penn Action~\cite{zhang2013actemes} contains videos of people performing sport actions in front of a camera. 3DPW~\cite{von2018recovering} is a recent video dataset with 3D ground truth of people outdoors obtained via IMUs. While these datasets are approaching ``in the wild'' settings outside a motion capture lab, they are still designed for performance capture where viewpoints are biased towards those that depict the entire body of people. This is in contrast to how people are depicted in films~\cite{gu2018ava,huang2020movienet} and other edited media~\cite{rockwell2020full}, which are abound with shot changes and close-up shots of people that induce a large truncation. We propose both optimization and direct prediction methods that can address these challenges and introduce MS-AVA in goals of further advancing 3D human mesh recovery from video.  

%% file: 04_multi_shot.tex
Here we present the first step of our workflow based on multi-shot optimization.  First, we describe the necessary preprocessing steps and the multi-shot optimization routine we use for pseudo ground truth generation. Then, we provide more details about the MS-AVA dataset we introduce.

\subsection{Preprocessing steps}
To apply our multi-shot optimization on a general video, we need a sequence of an individual within a scene, including shot changes. First, we detect 2D body joints using an off-the-shelf 2D pose tracker like OpenPose~\cite{cao2018openpose} or AlphaPose~\cite{fang2017rmpe}. While these methods obtain quite reliable 2D joint tracklets, they  fail across shot boundaries. To extend tracklet duration, we run a shot detection algorithm~\cite{sidiropoulos2011temporal,rao2020local}, and use a person re-identification network trained on movie data~\cite{huang2018person} to link identities across shots. The result is long 2D joint tracklets,  extending beyond shot boundaries, which are used as inputs to the multi-shot optimization.

\subsection{Multi-shot optimization}\label{sec:multishot}
Relying on the insight that the input shots depict a single underlying 4D scene, we propose a multi-shot optimization method that recovers a reliable 3D human mesh across shot changes. To make this more concrete,  Let us consider the case where we have access to two consecutive frames $t$ and $t+1$, before and after the shot boundary respectively. In the SMPLify~\cite{bogo2016keep} setting, we have data terms $E_{\text{proj}}^{t}$ and prior terms $E_{\text{prior}}^{t}$ for both frames. In the case of sequence input, relevant methods~\cite{arnab2019exploiting,kocabas2020vibe} include smoothness regularization on the \textit{camera frame}. Since the shot (and therefore the viewpoint) has changed, this is not applicable here. As a result, these methods would consider the shot change as a discontinuity and reconstruct each frame independently. 

In contrast to that, to leverage the continuity of the 3D body structure, we apply smoothness regularization in the \textit{canonical frame}. More specifically, we explicitly decompose the pose parameters $\theta$ 
to global orientation~$R_{\textrm{gl}}$ and body pose parameters $\theta_{\textrm{b}}$. By undoing the global orientation, we can compute the body joints $X_{\textrm{can}}=R_{\textrm{gl}}^T X$ in the canonical space. This formulation allows factoring out the camera motion, which can be abrupt, and imposing the smoothness term only in the canonical frame:
\begin{align}
%E_{\text{sm}} =  ||{J}^t_{can} - {J}^{t+1}_{can}||^2_2 + ||{\theta}^t_{b} - {\theta}^{t+1}_{b}||^2_2.
E^t_{\text{sm joint}} &=  ||{X}^t_{\textrm{can}} - {X}^{t+1}_{\textrm{can}}||^2_2\label{eq:sm1} \\
E^t_{\text{sm param}} &=  ||{\theta}^t_{\textrm{b}} - {\theta}^{t+1}_{\textrm{b}}||^2_2\label{eq:sm2}.
\end{align}
The sum of objectives is optimized over the entire sequence of length $T$:
\begin{align}
E = \sum_{t=1}^T (E_{\text{proj}}^{t} + E_{\text{prior}}^{t}) + \sum_{t=1}^{T-1} (E^t_{\text{sm joint}} + E^i_{\text{sm param}}),
\end{align}
returning model parameters $\Theta^t$ for every frame $t$ of the sequence. For faster convergence to a more accurate solution, we initialize our reconstruction with pose and shape estimates provided by a regression network~\cite{kolotouros2019learning}.

\subsection{Multi-Shot AVA dataset}\label{sec:msava}
Although the above workflow is applicable in many occasions with videos from TV series or movies, in this work, we focus primarily on the AVA dataset~\cite{gu2018ava}. AVA contains 300 movies annotated with human bounding boxes and atomic actions. Bounding box annotations are available at 1fps and organized in short tracklets. We also process the data at 1fps, but follow our preprocessing to extend the tracklet duration (i.e., link short tracklet of the same identity). Each tracklet is reconstructed in 3D with our multi-shot optimization (section~\ref{sec:multishot}). We call this new dataset \textbf{Multi Shot AVA},  or \textbf{MS-AVA}.

\begin{figure}[!ht]
\centering
\includegraphics[width=0.45\textwidth,trim={0 13cm 23.5cm 0},clip]{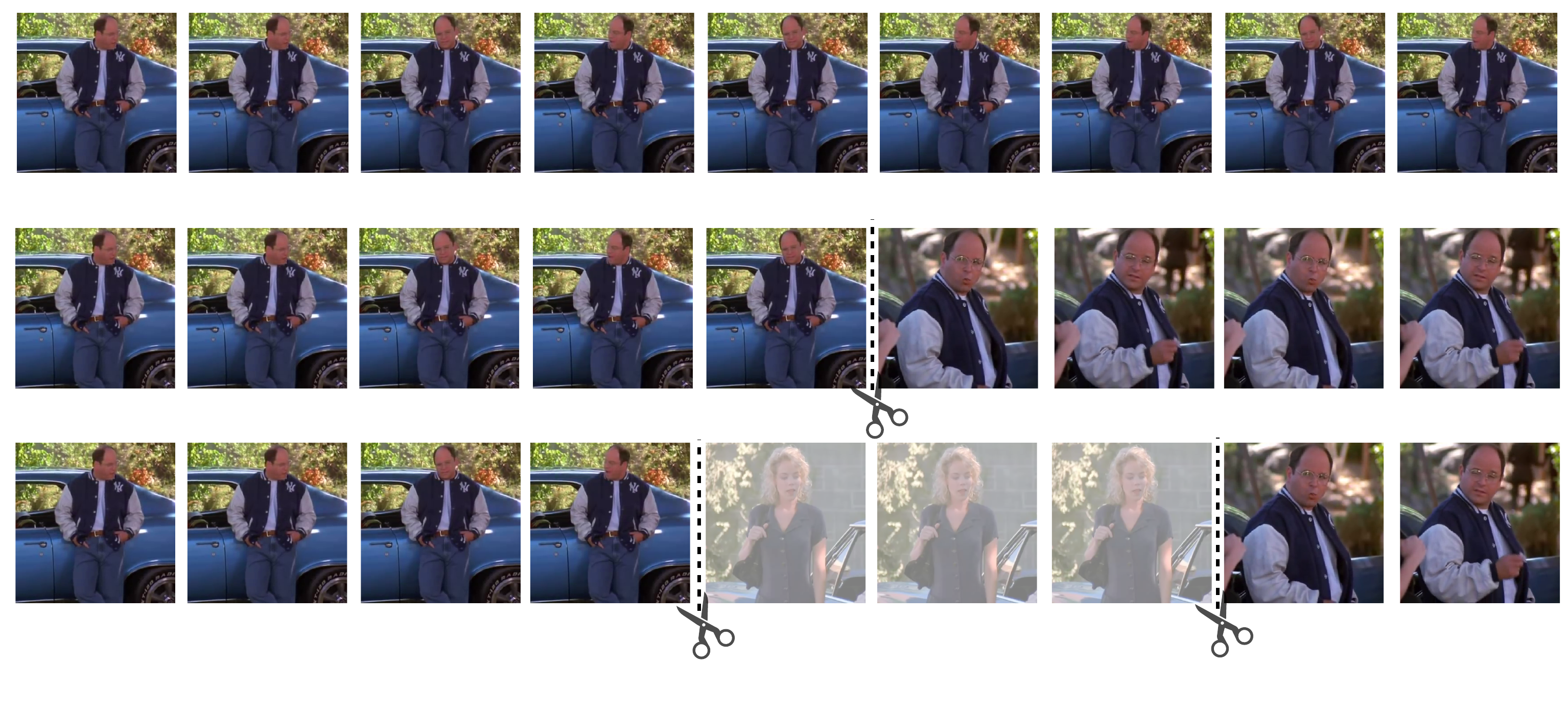}
%\vspace{-7mm}
\vspace{-5mm}
\caption{\textbf{Different type of sequences in MS-AVA.} The top row illustrates the most common type of sequence used in prior work with continuous sequence of an individual without shot changes. Movies include shot changes (middle) and show changes with non-continuous identity (bottom). Our approach can handle all three settings.}
  %\vspace{-2mm}
  \label{fig:data}
\end{figure}

Two important features of MS-AVA are the diverse and challenging visual conditions (\eg truncation), and the length and quantity of the sequences that it includes. In contrast to previous datasets~\cite{ionescu2013human3,von2018recovering,zhang2013actemes}, with MS-AVA, we are able to recover \textit{many}, \textit{long} and \textit{diverse} tracklets. There are two factors that help us achieve that: a) connecting tracklets across shots and b) keeping track of person identities, even if they are missing for some frames. Both aspects allow us to connect smaller, potentially overfragmented subsequences into longer sequences, useful for training temporal models. This statistic is also highlighted in Table~\ref{tab:msava}. More specifically, compared to single shot processing (``AVA single-shot''), we can increase the number of longer tracklets, if we consider identities that appear consecutively across shots (``AVA continuous identity''), or by re-identification after absence in some frames (``MS-AVA''). Examples of such sequences are illustrated in Figure~\ref{fig:data}.

\begin{table}
\centering
\resizebox{\columnwidth}{!}{
\small
\hspace{-3mm}
\begin{tabular}{@{}lll@{}}
\toprule
\multirow{2}{*}{Dataset} & Length (hours) & \#Tracklets \\
\cmidrule{2-2} \cmidrule{3-3}
& All\hspace{0.5em}/\hspace{0.5em}Long & All\hspace{0.5em}/\hspace{0.5em}Long \\
\midrule
Human3.6M~\cite{ionescu2013human3} & 6.45\hspace{0.5em}/\hspace{0.5em}6.45 & 330\hspace{0.5em}/\hspace{0.5em}330 \\
3DPW~\cite{von2018recovering} & 0.68\hspace{0.5em}/\hspace{0.5em}0.56 & 87\hspace{0.5em}/\hspace{0.5em}58 \\
Penn Action~\cite{zhang2013actemes} & 0.85\hspace{0.5em}/\hspace{0.5em}0 & 2.3k\hspace{0.5em}/\hspace{0.5em}0 \\
\midrule
AVA single-shot & 75.1\hspace{0.5em}/\hspace{0.5em}11.1 & 5.1k\hspace{0.5em}/\hspace{0.5em}1.3k \\
AVA continuous identity & 78.4\hspace{0.5em}/\hspace{0.5em}17.1 & 6.3k\hspace{0.5em}/\hspace{0.5em}1.8k \\
MS-AVA & 81.1\hspace{0.5em}/\hspace{0.5em}20.2 & 6.7k\hspace{0.5em}/\hspace{0.5em}2.1k \\
\bottomrule
\end{tabular}
}
\caption{\textbf{Raw duration and sequences for video datasets
}
Numbers are for a) all tracklets and b) long ones (length greater than 20 secs). MS-AVA provides many long sequences which is supported by considering multi-shot continuity and re-identification across disjoint subsequences (see also Figure~\ref{fig:data}).}
\label{tab:msava}
%\vspace{-2mm}
\end{table}

\begin{figure}[!ht]
\centering
\includegraphics[width=0.45\textwidth,trim={0 4cm 18cm 0},clip]{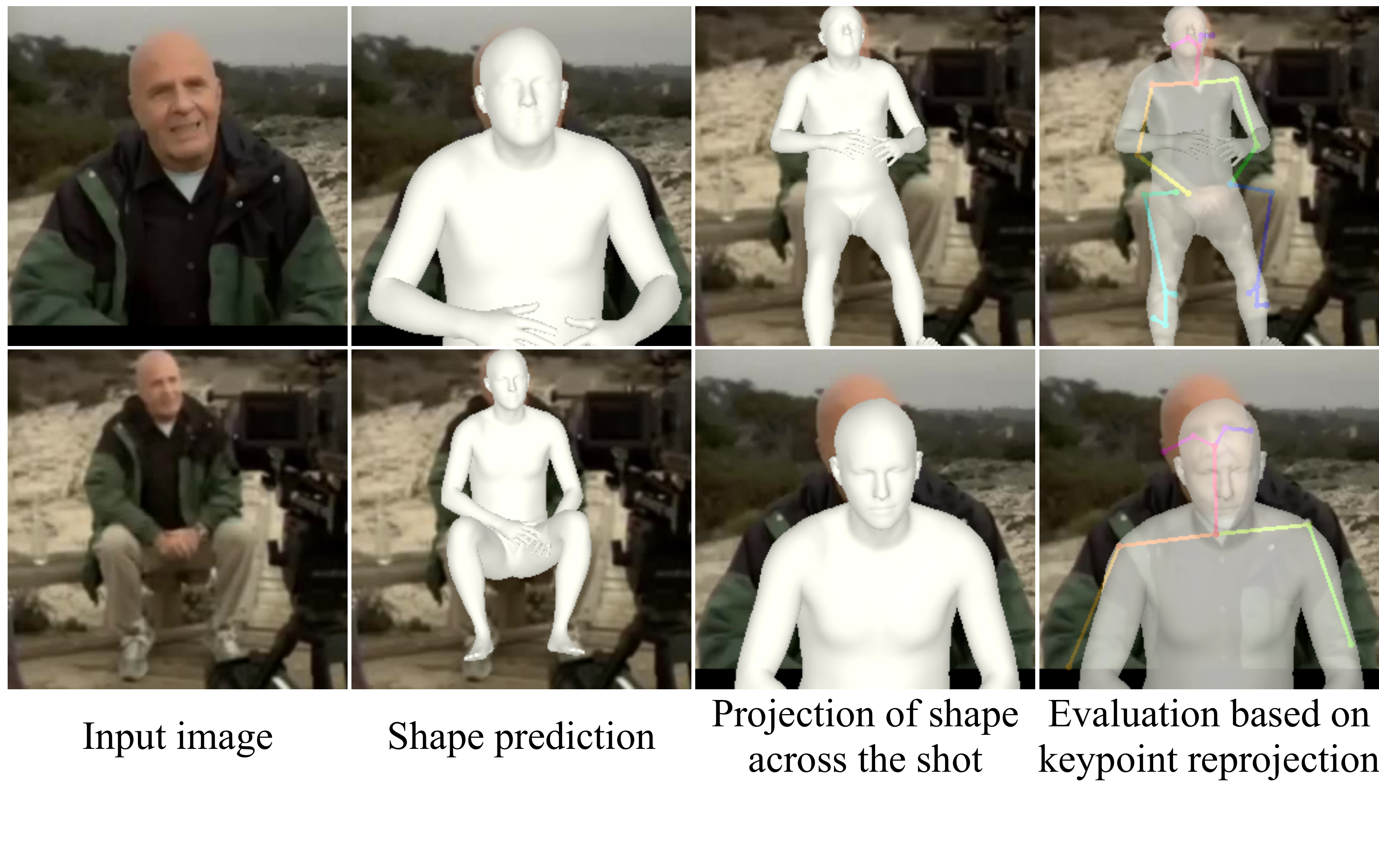}
\caption{\textbf{Novel view evaluation with \textit{cross-shot} PCK.} Given the shape prediction for frame $t$ (before the shot change), we project it to frame $t+1$ (after the shot change), and vice versa. We assess the 3D quality of the estimated shape by computing 2D reprojection metrics on this novel view.}
%\vspace{-.5em}
\label{fig:cross-shot-eval}
\end{figure}

Finally, our insight that pose changes smoothly across the shot boundary also offers the opportunity to evaluate the 3D accuracy of the recovered human mesh from monocular camera sequences via \textit{novel view evaluation}. Specifically, given a shot change from frame $t$ to frame $t+1$, we project the shape of frame $t$ to frame $t+1$, and vice versa. See Figure~\ref{fig:cross-shot-eval} for illustration. This allows us to evaluate the predicted shape using 2D reprojection metrics, \eg, PCK~\cite{yang2012articulated}. We refer to this metric as \textit{cross-shot PCK} and we use it to evaluate 3D shape quality in AVA, where only 2D keypoints are present.

\begin{figure*}[!ht]
\centering
\includegraphics[width=\textwidth,trim={0 10.5cm 12cm 0},clip]{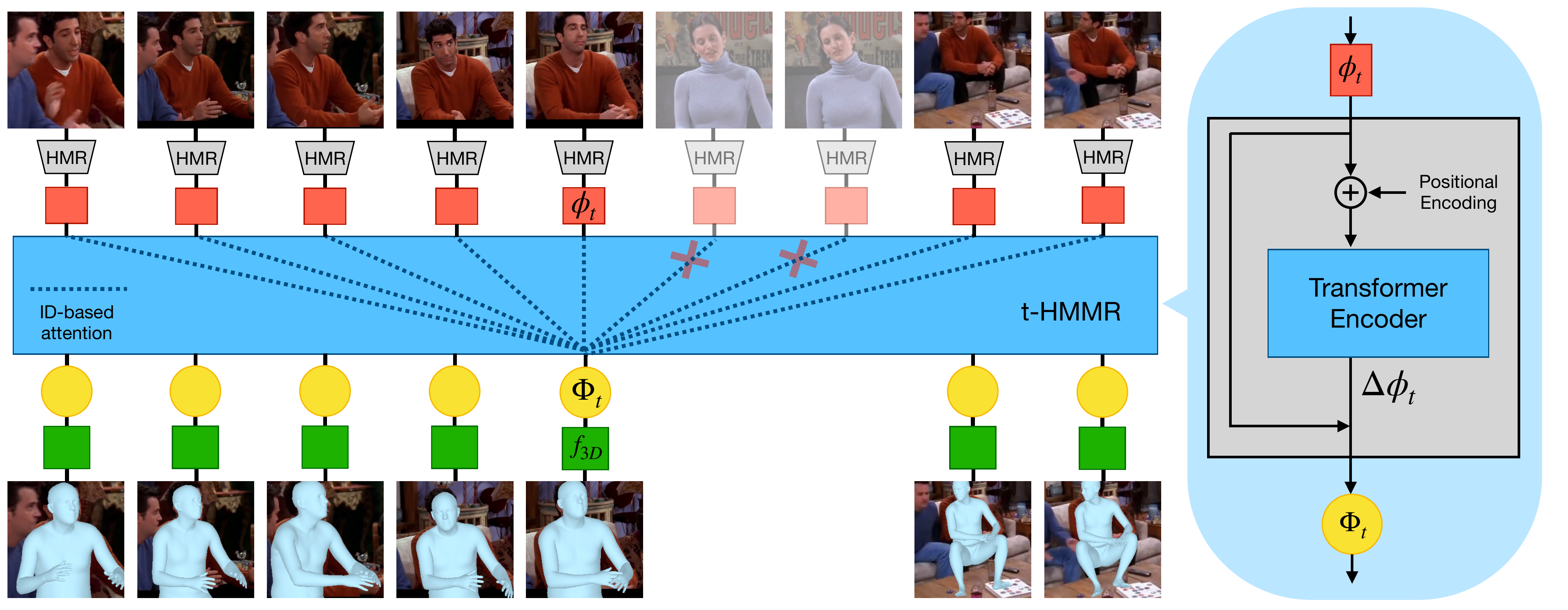}
\caption{\textbf{Architecture of t-HMMR:} To most effectively leverage the plethora of 3D pose sequences recovered from our data, we propose t-HMMR, a human mesh and motion recovery model based on the transformer architecture. Even if the identity of interest is not present in some frames, we can still benefit from the large temporal context, by setting the attention to zero for the invalid frames, while aggregating information from the relevant input images.}
\vspace{-.5em}
\label{fig:pipeline}
\end{figure*}

%% file: 05_human_mesh_recovery.tex
The 3D motion sequences we recovered with the offline multi-shot optimization step offer a rich source of data with pseudo ground truth 3D bodies. Here, we demonstrate how to incorporate this data in the training of direct prediction models for Human Mesh Recovery from single images or video, without the reliance on keypoint detections.

\subsection{Single-frame model}
The first step is to train an updated single-frame model. In general, the setting is similar to the original HMR~\cite{kanazawa2018end}. Let our image encoder for frame $I$ predict model parameters $\hat{\Theta}$ and camera parameters $\hat{\Pi}$. Model joints are projected to 2D locations $\hat{x}$. Our supervision for the network comes from the output of the multi-shot optimization for the corresponding frame, $\Theta_{\textrm{gt}}$, and the detected 2D joints $x_{\textrm{gt}}$.
\begin{align}
L_{\text{2D}} &=  ||\hat{x} - x_{\textrm{gt}}||_1 \label{eq:loss1} \\
L_{\text{smpl}} &=  ||\hat{\Theta} - \Theta_{\textrm{gt}} ||^2_2\label{eq:loss2}.
\end{align}
Our experiments show that the diversity and the challenging visual conditions (\eg, truncation) of MS-AVA, help improve the robustness of our single-frame model.

\subsection{Temporal model}
Using an updated and robust HMR, we proceed towards learning the temporal encoding function $f_{\textrm{movie}}$. In the past, this function has been represented by convolutional~\cite{kanazawa2019learning}, recurrent~\cite{kocabas2020vibe} or hybrid encoders~\cite{sun2019human}. However, in the above cases, the temporal training data come from curated collections of clean videos with continuous person tracking~\cite{ionescu2013human3,mehta2017monocular,zhang2013actemes}. In contrast, in more general use cases, including MS-AVA, video data can be more challenging with issues like shot changes or frames where the person of interest is not present (due to occlusion, tracking failures or re-identification failures). These cases are not easily handled by convolutional or recurrent encoders, which would require padding the inputs with zeros or concatenating all valid frames together (i.e., ignoring the time difference between consecutive frames).

To address these limitations,  we propose t-HMMR, a temporal model based on a pure transformer architecture~\cite{vaswani2017attention}. Transformers include an attention mechanism, allowing us to explicitly select the elements of the input sequence they will attend to. This is a convenient feature, particularly with the discontinuous nature of MS-AVA sequences (see Figure~\ref{fig:data}). Only considering the sequences where a person appears continuously without identity changes, limits the available training sequences (\eg, Table~\ref{tab:msava} for MS-AVA). The proposed transformer model has the advantage to address this challenge elegantly, making effective use of all the available sequences.

Our transformer encoder takes as input an intermediate HMR embedding $\{\phi_t\}$ of sequence of frames $\{I_t\}$. This sequence comes with a scalar value per-frame $\{v_t\}$, which indicates whether the person is present in frame $t$ ($v_t=1$), or not ($v_t=0$). A fixed positional encoding $p_t$ is added to the input features to indicate the time instance $t$ of each input element. The updated features are then processed by a \textit{transformer encoder layer}. This follows the architecture of the original transformer model, including a self-attention mechanism and a shallow feedforward network. The values $v_t$ are used to ensure that the invalid input frames will not contribute in the self-attention computation. The output of the transformer layer is a residual value $\Delta\phi_t$ added to the feature $\phi_t$ through a residual connection. The result of this block is the video feature representation $\Phi_t$. This is illustrated in Figure~\ref{fig:pipeline}. 

For training the transformer encoder, following prior work~\cite{kanazawa2019learning,kocabas2020vibe}, we fix the weights of the image encoder $f_{\textrm{im}}$, and only update the temporal encoder $f_{\textrm{movie}}$ and the parameter regressor $f_{\textrm{3D}}$. Similarly to the single-frame model, supervision is provided by the multi-shot optimization results, where we have corresponding losses with Equations~\ref{eq:loss1} and~\ref{eq:loss2}, $L^t_{\text{2D}}$ and $L^t_{\text{smpl}}$ respectively, for each frame $t$. Also, to further encourage temporal consistency, smoothness losses are applied on 3D joints $L^t_{\text{sm joint}}$ and 3D model parameters $L^t_{\text{sm joint}}$ (equivalent to equations~\ref{eq:sm1} and~\ref{eq:sm2} respectively).

%% file: 06_experiments.tex
The focus of our quantitative evaluation is three-fold: we first demonstrate the quality of our multi-shot optimization results; then we investigate the effect of our data for improving the performance and robustness of a single-frame human mesh recovery model; finally we examine our temporal model and validate the importance of leveraging the transformer architecture. In the next paragraph, we present in detail each one of these aspects.

\subsection{Implementation details}
For our HMR baseline, we retrain an HMR model on the standard datasets (Human3.6M~\cite{ionescu2013human3}, COCO~\cite{lin2014microsoft}, MPII~\cite{andriluka20142d}) using pseudo ground truth SMPL parameters from SPIN~\cite{kolotouros2019learning}, and incorporating the cropping augmentation scheme~\cite{joo2018total,rockwell2020full}. We use this baseline to initialize our multi-shot optimization and for ablative experiments. After the offline multi-shot optimization, our final HMR model is trained with the same strategy, including the data from MS-AVA. For the t-HMMR model, we freeze the image encoder of HMR, as done in~\cite{kanazawa2019learning,kocabas2020vibe}, for computational reasons, and train the temporal encoder and 3D regressor only.

\subsection{Quantitative evaluation}

\begin{figure}[h]
\centering
\includegraphics[width=0.45\textwidth]{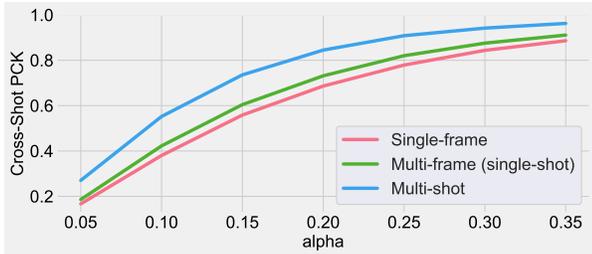}
\caption{\textbf{Multi-shot optimization evaluation on AVA.} We show cross-shot PCK for varying thresholds $\alpha$.  Our multi-shot optimization outperforms optimization baseline applied on a single-frame or single-shot (multiple frames that do not span shot changes).}
\vspace{-1em}
\label{fig:multishot-quant}
\end{figure}

\noindent\textbf{Multi-shot optimization:}
Our multi-shot optimization integrates information across the shot boundary to improve pose reconstruction. To evaluate its success, we use the cross-shot PCK metric as discussed in Section~\ref{sec:msava}. As a sanity check, we compare with two optimization based baselines, one that considers a single frame only~\cite{bogo2016keep}, and one that considers a temporal sequence without shot changes~\cite{arnab2019exploiting,kocabas2020vibe}. The results are presented in Figure~\ref{fig:multishot-quant}, where multi-shot optimization outperforms the two baselines. As expected, the multi-shot optimization outperforms the two baselines, since the formulation of the objective function allows the integration of information from multiple shots, and the large gap in performance indicates that it is very successful in this regard. Qualitative examples of this behavior are presented in Figure~\ref{fig:multishot-qual} and in the Supplementary.

\begin{figure}[!ht]
\centering
\includegraphics[width=0.45\textwidth,trim={0 1.5cm 8cm 0},clip]{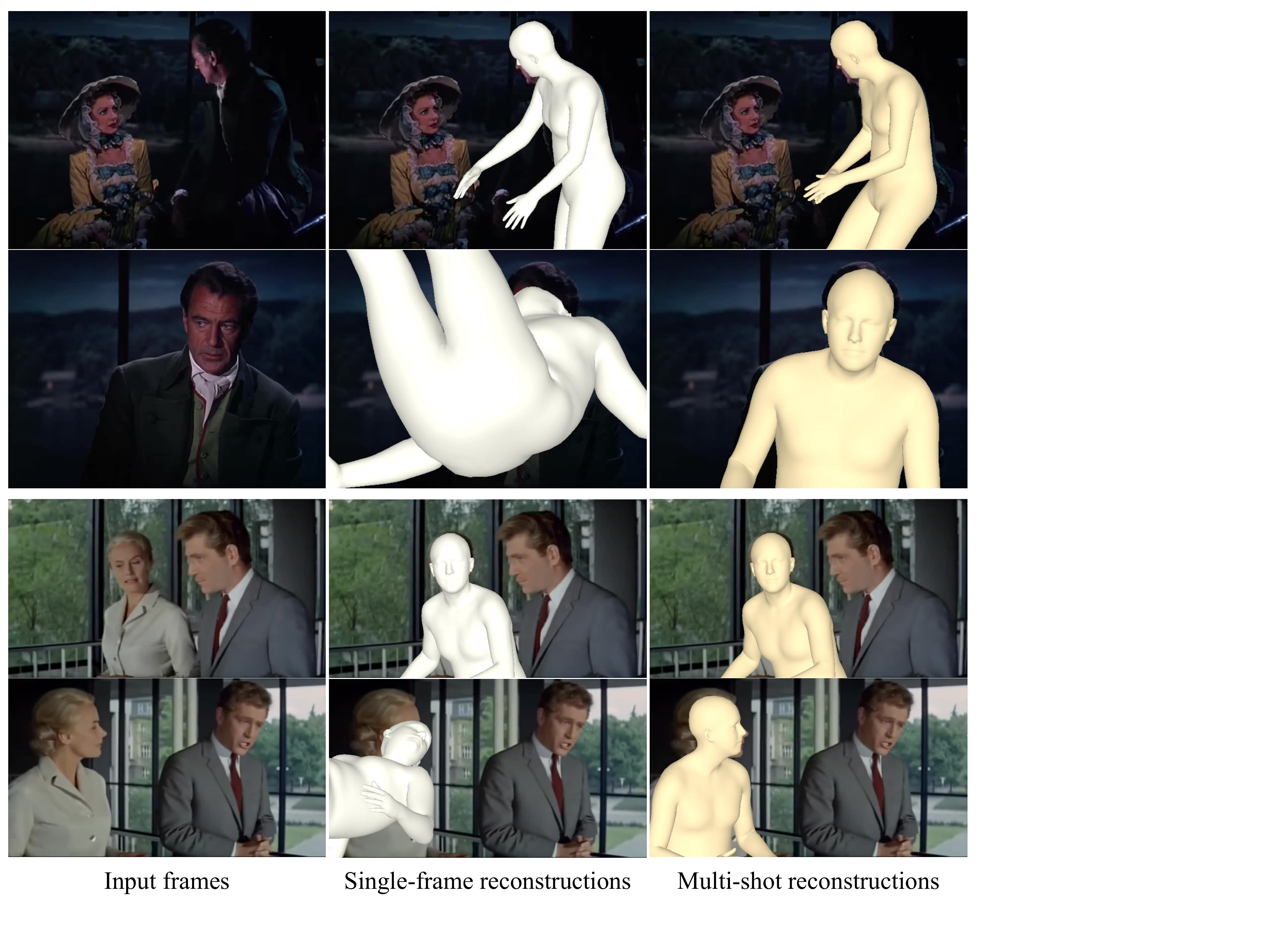}  
\caption{\textbf{Qualitative effect of our multi-shot optimization.} Although a single frame baseline fails on the more challenging frames with heavy truncation (center), our multi-shot optimization  leverages information from the less ambiguous frame across the  shot boundary to get a more accurate 3D reconstruction.}
%\vspace{-.5em}
\label{fig:multishot-qual}
\end{figure}

\begin{figure}[h]
  \centering
\includegraphics[width=0.45\textwidth,trim={0 1cm 5cm 0},clip]{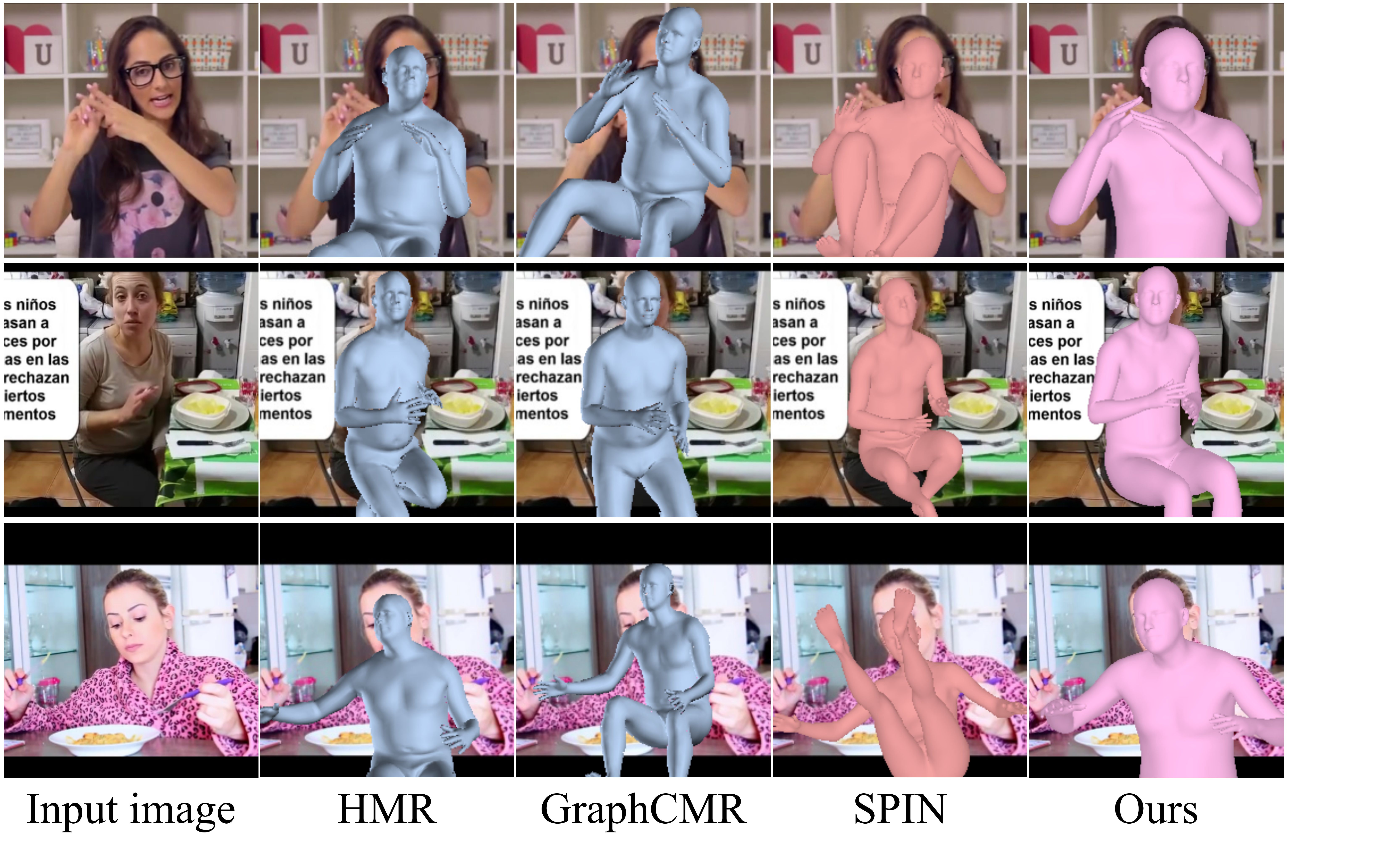} 
\caption{\textbf{Comparison against recent methods on Partial Humans dataset~\cite{rockwell2020full}.} Qualitative comparison with recent state of the art on the Partial Humans dataset~\cite{rockwell2020full}. Our model is significantly more robust in the presence of truncations.
}
% \vspace{-.5em}
\label{fig:partial}
\end{figure}

\noindent\textbf{Single-frame model:}
The introduced MS-AVA dataset provides a rich source of data for training our single-frame model for human mesh recovery. Considering the nature of the data, they could help improving the robustness of our model, particularly when it comes to truncation, a common mode of failure for the state-of-the-art models~\cite{rockwell2020full}. We demonstrate this on the Partial Humans dataset~\cite{rockwell2020full} in Table~\ref{tab:missing_partial}. Our model trained with the MS-AVA data outperforms the state-of-the-art approaches. Many of these methods are trained primarily with full body images so there are significant failures under truncation, whereas our network can be robust in these cases too. This robustness is illustrated in Figure~\ref{fig:partial}, where we provide qualitative results for different methods on the ``Uncropped'' subset of Partial Humans. Moreover, our HMR model trained on MS-AVA outperforms our baseline trained with synthetic cropping augmentation, underlining the effect of our MS-AVA data at achieving better robustness. Finally, we also evaluate on the 3DPW dataset~\cite{von2018recovering} in Table~\ref{tab:missing_3dpw}, where we compare with the most relevant approaches and observe the same trends.

\begin{table}
\centering
\small
\hspace{-3mm}
\tabcolsep=2.95mm
\begin{tabular}{@{}lcccc@{}}
\toprule
& \multicolumn{4}{c}{PCKh on Partial Humans Uncropped $\uparrow$}\\
\cmidrule{2-5}
Method & VLOG & YouCook & Instr & Cross \\
\midrule
HMR~\cite{kanazawa2018end} & 81.2 & 93.6 & 86.9 &92.7 \\
GraphCMR~\cite{kolotouros2019convolutional} & 65.7 & 80.1 & 77.5 & 79.3 \\
SPIN~\cite{kolotouros2019learning} & 73.4 & 85.1 & 85.6 & 85.5 \\
Partial Humans$^{*}$~\cite{rockwell2020full} & 68.7 & 95.4 & 77.9 & 91.1 \\
\midrule
HMR (retrained) & 86.9 & 96.8 & 92.4 & 96.2 \\
\hspace{3mm} + MS-AVA data & \bf{90.3} & \bf{98.9} & \bf{94.1} & \bf{98.2} \\
\bottomrule
\end{tabular}
\vspace{-2mm}
\caption{\textbf{PCKh results for different single-frame models on the uncropped subset of the Partial Humans dataset~\cite{rockwell2020full}.} Best performance is achieved when we train on MS-AVA data. Note $^{*}$ operates in the harder setting, which uses the entire image as an input while others operate on cropped bounding boxes. 
}
\label{tab:missing_partial}
\vspace{-.5em}
\end{table}

\begin{table}
\centering
\small
\hspace{-3mm}
\tabcolsep=2.95mm
\begin{tabular}{@{}lcc@{}}
\toprule
Method & PA-MPJPE\\
\midrule
HMR~\cite{kanazawa2018end} & 81.3 \\
SPIN~\cite{kolotouros2019convolutional} & 59.2 \\
\midrule
HMR (retrained) & 59.2 \\
\hspace{3mm} + MS-AVA data & \bf{57.8} \\
\bottomrule
\end{tabular}
%\vspace{-2mm}
\caption{\textbf{Single-frame regression evaluation on 3DPW.} The numbers are PA-MPJPE expressed in mm. Best performance is achieved when we include our MS-AVA data in training.}
\label{tab:missing_3dpw}
\vspace{-.7em}
\end{table}

\begin{figure}[h]
\centering
\includegraphics[width=0.45\textwidth]{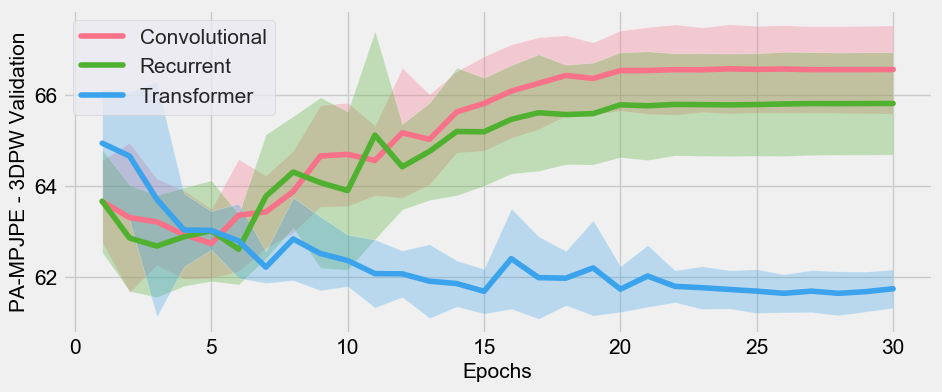}
\caption{\textbf{Comparison of different temporal encoders on the VIBE training setting~\cite{kocabas2020vibe}.} Performance is tracked over multiple runs. Although all encoders can achieve good performance at least once during the whole training schedule, transformer training is more stable and
converges consistently to a good model.
}
% \vspace{-.5em}
\label{fig:vibe}
\end{figure}

\noindent\textbf{Temporal model:}
For our temporal model, first we investigate the transformer encoder \textit{from an architecture viewpoint}, factoring out the effect of challenging/incomplete data. To this end, we compare it directly with other architecture choices for temporal encoders, i.e., the convolutional encoder of HMMR~\cite{kanazawa2019learning} and the recurrent encoder of VIBE~\cite{kocabas2020vibe}. For a direct comparison, we follow the exact training data/schedule with VIBE using their public implementation, and replacing only the temporal encoder. We perform multiple training runs and plot validation performance in Figure~\ref{fig:vibe}. If we consider the best performance across all iterations, all models achieve similar results. However, the convolutional and recurrent encoders tend to diverge after a few epochs, while the transformer encoder is more stable. Simultaneously, this version still achieves state-of-the-art results (PA-MPJPE of 56.1mm vs 56.5mm for VIBE on the 3DPW test set), even though it is trained on less data than the full VIBE model.

Next, we also demonstrate the suitability of the transformer model in cases of \textit{learning and testing on incomplete data}, \eg, on movies. In these cases, the architectures other than the transformers do not have a straightforward way to handle missing frames. We experimented with different alternatives, and we found that padding missing frames with zero features performed the best for the convolutional and the recurrent encoder. Using this strategy, we report cross-shot PCK results on AVA for different encoder choices in Table~\ref{tab:missing}. In the first setting, we test on the clean sequences (``test: clean'' - consecutive sequences without missing frames), trained using our clean sequences (``train: clean'') from AVA. In this setting all models perform well, however, when we add the sequences with missing frames in the training (``train: all''), the transformer model benefits the most from the additional training data. In the second setting, we test on all sequences (``test: all''). The transformer still performs the best in this more challenging case.

\begin{figure}[!ht]
\centering
\includegraphics[width=0.45\textwidth,trim={0 2cm 23cm 0},clip]{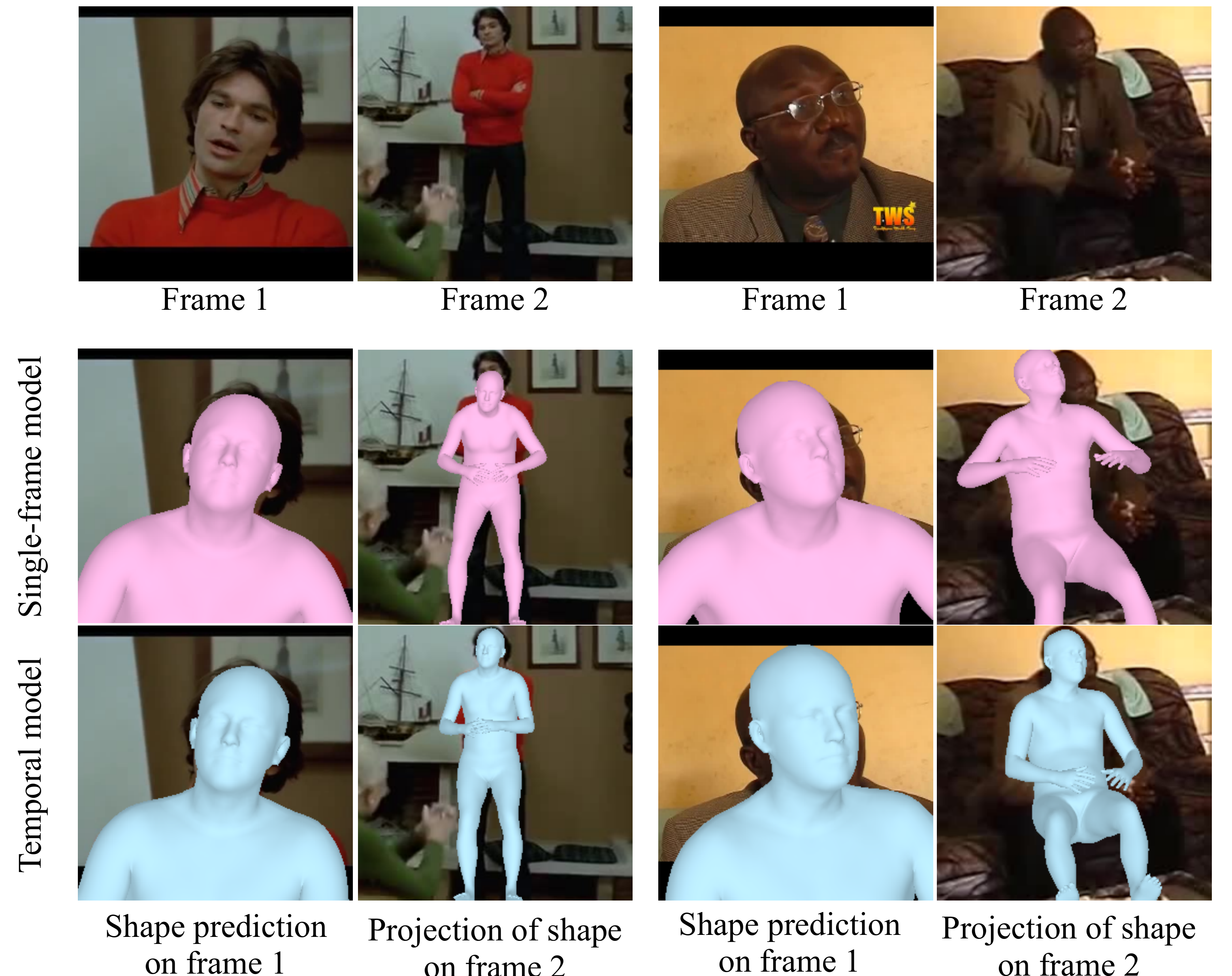}
\caption{\textbf{Effect of t-HMMR model.} While the single-frame HMR prediction for frame~1 can be inconsistent with frame~2, our temporal t-HMMR model successfully integrates information over a temporal window and estimates a body pose for frame 1 that is consistent with frame 2.
}
\vspace{-1em}
\label{fig:temporal}
\end{figure}

Finally, in Figure~\ref{fig:temporal} we provide some example reconstructions of our temporal t-HMMR model, in comparison with the single frame HMR model, both trained on MS-AVA. While our HMR obtains reasonable results, output from t-HMMR is more consistent as it has a larger temporal context.

\begin{table}
\centering
\small
\hspace{-3mm}
\tabcolsep=2.95mm
\begin{tabular}{@{}l|cl|c@{}}
\toprule
Temporal & train: clean & train: all & train: all \\
Model & \multicolumn{2}{c|}{test: clean} & test: all \\
\midrule
Convolutional & 68.8 & 69.7 & 67.5 \\
Recurrent & 70.8 & 70.8 & 68.5 \\
Transformer & 70.7 & 73.8 & 70.7 \\
\bottomrule
\end{tabular}
%\vspace{-2mm}
\caption{\textbf{Multi-frame evaluation on AVA, for different training/testing protocols.} The numbers are cross-shot PCK values. Transformers enjoy the largest benefit by incorporating sequences with missing data in the \textit{training}, while they also perform the best when \textit{tested} on sequences with missing data.}
\label{tab:missing}
%\vspace{-2mm}
\vspace{-1em}
\end{table}

%% file: 07_conclusion.tex
We introduce a new task of 3D human reconstruction from multiple shots, proposing a framework to generate training data from edited media like movies, and use them in different human mesh recovery tasks. Our contributions span all three major forms of Human Mesh Recovery: off-line iterative optimization, single-view prediction, and temporal prediction. Our experiments demonstrate the significance of our contributions in each of these aspects. Future work can investigate further the available data from MS-AVA for other downstream applications or apply our data generation framework in other edited media.

\footnotesize
\noindent
{\bf Acknowledgements:} This research was supported by BAIR sponsors.